\documentclass[letterpaper]{article} 
\usepackage{aaai24}  
\usepackage{times}  
\usepackage{helvet}  
\usepackage{courier}  
\usepackage[hyphens]{url}  
\usepackage{graphicx} 
\usepackage{natbib}  
\usepackage{caption} 
\usepackage{algorithm}
\usepackage{algorithmic}

\usepackage{newfloat}
\usepackage{listings}
\DeclareCaptionStyle{ruled}{labelfont=normalfont,labelsep=colon,strut=off} 
\floatstyle{ruled}
\newfloat{listing}{tb}{lst}{}
\floatname{listing}{Listing}
\title{Diverse and Fine-Grained Instruction-Following Ability Exploration with Synthetic Data}
\author{
    Zihui Gu\textsuperscript{\rm 1,2}\equalcontrib, Xingwu Sun\textsuperscript{\rm 2,3}\equalcontrib, Fengzong Lian\textsuperscript{\rm 2}, Zhanhui Kang\textsuperscript{\rm 2}, Cheng-Zhong Xu\textsuperscript{\rm 3}, Ju Fan\textsuperscript{\rm 1}\thanks{Ju Fan is the corresponding author.}\\
}
\affiliations{
    \textsuperscript{\rm 1}Renmin Univerisity of China\\
    \textsuperscript{\rm 2}Tencent Inc.\\
    \textsuperscript{\rm 3}University of Macau\\
    \{guzh, fanj\}@ruc.edu.cn, sunxingwu01@gmail.com, \{faxonlian, kegokang\}@tencent.com, czxu@um.edu.mo
    
}

\usepackage{bibentry}
\usepackage{xspace}
\usepackage{graphicx}
\usepackage{textcomp}
\usepackage{xcolor}
\usepackage{booktabs}
\usepackage{multirow}
\usepackage{amsmath}
\usepackage{xspace}
\usepackage{enumitem}
\usepackage{mathrsfs}
\usepackage{makecell}
\usepackage{xparse}
\usepackage{url}
\usepackage{pifont}
\usepackage{bbding}
\usepackage{multirow}
\usepackage{subfigure}
\usepackage{xcolor,listings}

\usepackage{color, colortbl}
\usepackage{verbatim}

\newlist{compactitem}{itemize}{3} 
\setlist[compactitem]{label=\textbullet, nosep, leftmargin=0cm,itemindent=.5cm}

\definecolor{dkgreen}{rgb}{0,0.6,0}
\definecolor{gray}{rgb}{0.5,0.5,0.5}
\definecolor{mauve}{rgb}{0.58,0,0.82}
\definecolor{green1}{HTML}{98dfaf}
\definecolor{green2}{HTML}{98c39a}
\definecolor{green3}{HTML}{516850}
\newcommand{\dataset}{\textsc{DINGO}\xspace}

\newcommand{\llm}{\textsc{LLM}\xspace}
\newcommand{\llms}{\textsc{LLM}s\xspace}

\newcommand{\bi}{\begin{compactitem}}
\newcommand{\ei}{\end{compactitem}}

\newcommand{\be}{\begin{enumerate}}
\newcommand{\ee}{\end{enumerate}}
\newcommand{\beqn}{\begin{eqnarray*}}
\newcommand{\eeqn}{\end{eqnarray*}}

\newcommand{\ie}{{\em i.e.,}\xspace}
\newcommand{\eg}{{\em e.g.,}\xspace}

\begin{document}
\maketitle
\begin{abstract}
Instruction-following is particularly crucial for large language models (\llms) to support diverse user requests. While existing work has made progress in aligning \llms with human preferences, evaluating their capabilities on instruction-following remains a challenge due to complexity and diversity of real-world user instructions. While existing evaluation methods focus on general skills, they suffer from two main shortcomings, \ie lack of fine-grained task-level evaluation and reliance on singular instruction expression. To address these problems, this paper introduces \dataset, a fine-grained and diverse instruction-following evaluation dataset that has two main advantages: (1) \dataset is based on a manual-annotated, fine-grained and multi-level category tree with 130 nodes derived from real-world user requests; (2) \dataset includes diverse instructions, generated by both GPT-4 and human experts. Through extensive experiments, we demonstrate that \dataset can not only provide more challenging and comprehensive evaluation for \llms, but also provide task-level fine-grained directions to further improve \llms.
\end{abstract}
\section{Introduction}
\label{sec:introduction}

Recently, Large language models (\llms) exhibit surprising capabilities not previously seen in smaller models, which are often referred to as {\it emergent abilities}~\cite{DBLP:journals/tmlr/WeiTBRZBYBZMCHVLDF22}, including {\it in-context learning}, {\it chain-of-thought}, and {\it instruction-following} abilities. Among them, the {\it instruction-following} ability is crucial to the interaction between humans and \llms (\eg ChatGPT). Existing studies~\cite{gpt4, vicuna, DBLP:journals/corr/abs-2306-04751,flanv2} align \llms with human instructions using supervised instruction-tuning or reinforcement learning from human feedback (RLHF), which enables \llms to understand human instructions and make high-quality responses. Nonetheless, due to the complexity and diversity of human instructions, it remains a challenge to comprehensively evaluate the {\it instruction-following} ability of \llms.

\begin{figure}[t!]
  \centering
  \includegraphics[width=1.\linewidth]{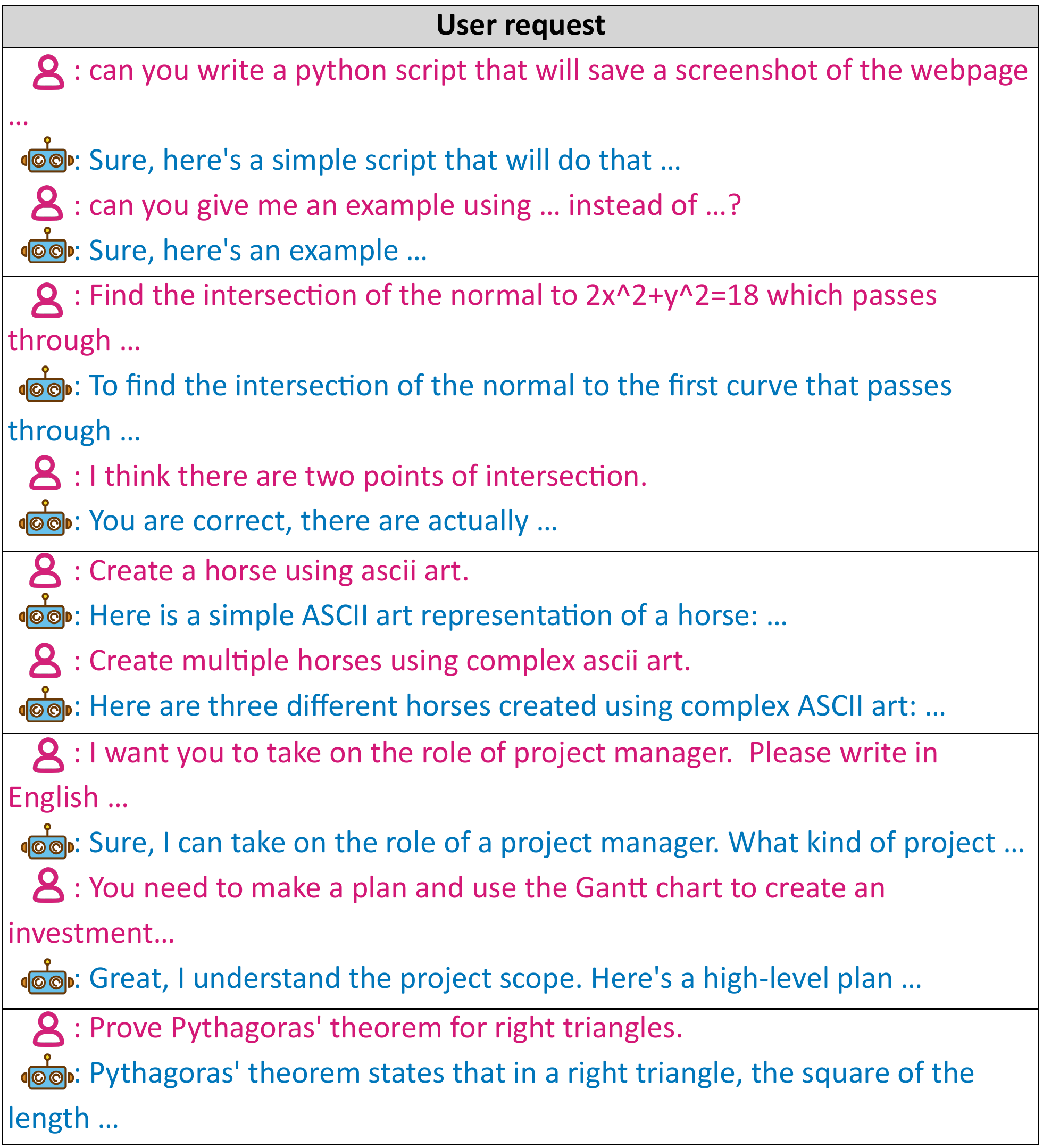}
\caption{Different user request examples extracted from ShareGPT.}
  \label{fig:sharegpt_examples}
\end{figure}
Existing studies evaluate the {\it instruction-following} ability from the perspective of general skills. For example, InstructEval~\cite{chia2023instructeval} assesses \llm's instruction-following ability based on three general abilities: problem-solving, writing, and alignment to human values.
Flask~\cite{DBLP:journals/corr/abs-2307-10928} shifts the original coarse-grained scoring process to instance-wise skill scoring setup, and defines 4 primary abilities, divided into 12 specific skills, to assess the performance of \llms. However, there are still two shortcomings in existing evaluation methods: 

\bi
\item The {\bf lack of fine-grained task-level evaluation} poses challenges in improving the instruction-following ability of \llms. For example, the {\tt Factuality} skill used in FLASK~\cite{DBLP:journals/corr/abs-2307-10928} includes many sub-tasks such as ``History Knowledge QA'' and ``Chemical Knowledge QA''. 
Consequently, even if we recognize that a particular \llm is deficient in this skill, it is challenging to pinpoint the exact aspects of the instruction-following ability that the \llm needs to be improved.
Specifically, if the performance of the \llm is not satisfactory in ``Chemical Knowledge QA'', it is not clear whether this is because the \llm's response contains non-standard chemical formulas.
Similarly, if the \llm cannot perform well in ``History Knowledge QA'', it could potentially be because the key points are not clearly outlined in the \llm's response. 

\item The {\bf expression of instructions tends to be singular}, resulting in a gap between real-world user instructions and existing evaluation datasets. Existing datasets~\cite{chia2023instructeval} often use previous NLP datasets as evaluation data for specific skills, such as employing DROP~\cite{Dua2019DROPAR} to evaluate the {\tt Comprehension} ability, and design a specific instruction template for the dataset. However, in real-world scenarios, users express their requests in various ways. Figure~\ref{fig:sharegpt_examples} shows several examples extracted from the ShareGPT website, a platform where users voluntarily share their interaction records with \llms. As can be seen, the styles and attitudes of user instructions are very diverse: users may ask questions directly (\ie Concise) or set specific roles to ask questions (\ie Role-play). Therefore, it could be very beneficial to evaluate the \llm's instruction-following ability on these diverse instruction expressions.
\ei

To address the aforementioned shortcomings, in this paper, we present \dataset, a \underline{D}iverse and F\underline{i}ne-grained I\underline{n}struction-Followin\underline{g} evaluation dataset. First, to support fine-grained instruction-following evaluation, we manually annotate a multi-level category tree with {130} nodes and 4 levels, based on the user instructions extracted from ShareGPT. This category tree encompasses tasks that users would want \llms to complete in real-world scenarios, making it highly practical. Equipped with its multi-level structure, the category tree supports analyzing instruction-following ability at different granularities, and thus can address the shortcomings of \llm at task-level. Second, we prepare diverse instruction data for each category to comprehensively examine the instruction-following ability. Considering that user requests on ShareGPT have been used for instruction-tuning in many \llms, such as vicuna~\cite{vicuna} and TÜLU~\cite{DBLP:journals/corr/abs-2306-04751}, we avoid data leakage by not directly using data from ShareGPT for evaluation. Instead, we employ GPT-4 to simulate various instruction styles, attitudes, and languages derived from ShareGPT, and generate diverse instruction data for each category. In addition, considering the weaknesses of \llms in mathematics and logical reasoning, we utilize existing human-annotated datasets (\eg GSM8K~\cite{gsm8k}) as basic questions and guide GPT-4 to generate diverse instructions from the basic questions to ensure the instruction quality. For example, a math question from GSM8K, {\it ``Ronnie was given 5 while Rissa was given thrice as much …''} would be transformed by GPT-4 into a role-playing instruction form: {\it ``Act as a patient math teacher to answer this question step by step: Ronnie was given 5 while Rissa was given thrice as much …''}. Based on the above methods, we, in total, collect {5026} diverse samples in \dataset to comprehensively evaluate the instruction-following ability of \llms. 

Based on \dataset, we conduct extensive experiments to evaluate instruction-following of 10 different \llms, and obtain the following findings.
(1) Even if an instruction-tuned \llm performs well on coarse-grained categories, its performance on fine-grained categories may be diversified and, sometimes, it could even be worse than the base \llm without instruction fine-tuning. (2) Our dataset with diverse instructions presents more significant challenges to \llms to generate responses that align with human preferences.

Our contributions can be summarized as follows:

\bi
\item We publicly release a multi-level task category tree consisting of 130 nodes, designed to support instruction-following evaluations at various granularities.

\item We collect 5026 diverse and high-quality instructions based on real-world user instructions, presenting more significant challenges for \llms in generating responses that align with human preferences.

\item We conduct a comprehensive evaluation on 10 representative \llms, and the experimental results demonstrate that \dataset can support more extensive and challenging evaluation on the instruction-following ability, as well as provide fine-grained guidance to further improve \llms. We release the \dataset dataset at Github\footnote{https://github.com/ruc-datalab/DINGO}.
\ei
\section{Background: Instruction-Following Ability of Large Language Models}
\label{sec:preliminaries}

Language models (LMs) are designed to comprehend and produce text that resembles human language (\eg BERT~\cite{bert}, GPT2~\cite{gpt2}). 
Recently, researchers have discovered that scaling LMs to 
large LMs (\llms) (\eg ChatGPT, GPT-4~\cite{gpt4}, LLaMA~\cite{llama}) by increasing the model size or amount of training data can significantly enhance their downstream task abilities. Moreover, the existing studies also show that \llms demonstrate surprising abilities that have not been seen in previous smaller LMs~\cite{agi_sparks,rae2021scaling,DBLP:conf/nips/BrownMRSKDNSSAA20}, such as {\it in-context learning} and {\it instruction-following}.

{\it Instruction-following} is an important ability for \llms to interact with real-world users. This means that the model can complete various tasks based on a wide range of natural language instructions provided by humans, including polishing articles (\eg {\it Polish this email above in very urgent tone: \{Email\}.}), solving math problems (\eg {\it I need to calculate how long my gas canister will last for 360 burener.}), providing travel plans (\eg {\it Plan a trip to Jindo for 2 nights and 3 days.}), etc. \llms can obtain the instruction-following ability in the following two ways: (1) supervised learning using instruction-following datasets (\eg vicuna~\cite{vicuna}), and (2) reinforcement learning from Human Feedback(\eg Llama2-chat~\cite{llama2}). 

In this work, we aim to evaluate the capabilities of existing \llms on instruction-following across a variety of tasks and various instruction expressions, and provide a comprehensive benchmark \dataset to promote in-depth analysis of the instruction-following ability of \llms.
\section{The \dataset Dataset}
\label{sec:dataset}
\begin{figure*}[h]
  \centering
  \includegraphics[width=1.\linewidth]{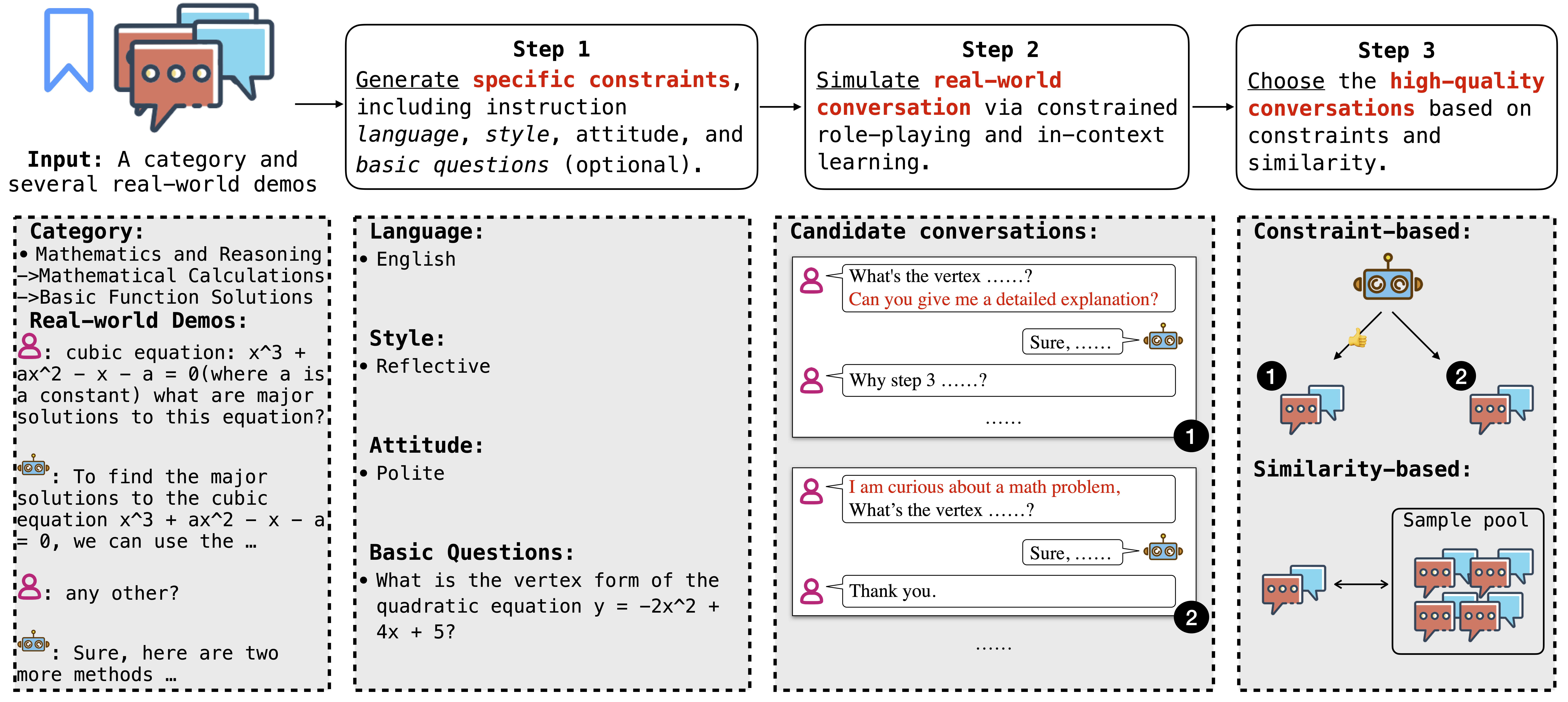}
\caption{A high-level pipeline of Instruction Data Collection.}
  \label{fig:pipeline}
\end{figure*}

Our goal is to generate a fine-grained {\tt category tree} and diverse {\tt instructions}. To achieve this goal, we first collect real-world user instructions as seed data. Then, we manually classify the seed data to obtain a fine-grained {\tt category tree}. Finally, based on seed data and {\tt category tree}, we collect diverse {\tt instructions} for each category by guiding GPT-4~\cite{gpt4} to simulate various instruction styles, attitudes, and languages.

\subsection{Seed Data Collection}
\label{sec:seed_data_collection}
To obtain real-world instruction-following data, we utilize public data from ShareGPT (https://sharegpt.com/), which is a platform for users to share their interactions with \llms (\eg GPT-4). Following previous work~\cite{vicuna,DBLP:journals/corr/abs-2306-04751}, we use the `html\_cleaned` version\footnote{https://huggingface.co/datasets/anon8231489123/\\ShareGPT\_Vicuna\_unfiltered/tree/main/HTML\_cleaned\_raw\_dataset} and truncate conversations with more than 2048 tokens. Based on this, we obtain 7265 seed samples from ShareGPT.

\subsection{Category Tree Annotation}
\label{sec:category_tree_annotation}

\begin{table}[h!]
    \centering
    \begin{tabular}{p{2cm}|p{5.5cm}}
         \hline
         {\bf First-level} & {\bf Second-level} \\
         \hline
         Language Understanding & Relationship Judgement; Classification; Sorting; Error Correction; Joke Explanation; Information Extraction \\
         \hline
         Code & Text2Code; Code2Text; Code2Code \\
         \hline
         Knowledge Unitilization & Open-book Questions; Close-book Questions \\
         \hline
         Creation & Thematic creation; Specialized writing; Plan; Non-verbal creation; Simulation creation \\
         \hline
         Language Generation & Question Generation; Rewriting/Paraphrasing; Summary/Abstract/Title; Translation \\
         \hline
         Mathematics and Reasoning & Word Problems; Mathematical theorem; Combinatorics; Mathematical Calculations; Common Sense Reasoning; Logical Reasoning \\
         \hline
    \end{tabular}
    \caption{The first and second level categories of \dataset.}
    \label{tab:category_tree}
\end{table}

\begin{table}[]
    \centering
    \begin{tabular}{p{8cm}}
         \hline
         {/* Task description */} \\
         I need you to simulate the conversation between ``human'' and ``AI''. I will specify some constraints, including \ldots \\
         {/* Demos from seed data */} \\
         {\bf Category}: Mathematics and Reasoning $\rightarrow$ Applied Problems; {\bf Language}: English; {\bf Style}: Concise; {\bf Attitude}: Command;\\
         {\bf Conversation}: Human: Calculate how long my gas \ldots ? AI: \ldots \\
         {\bf Category}: Mathematics and Reasoning $\rightarrow$ Combinatorics; {\bf Language}: English; {\bf Style}: Roly-play; {\bf Attitude}: Polite; \\
         {\bf Conversation}: Human: You are a math teacher, please explain this question step by step: There are two rows in a classroom \ldots ? AI: \ldots \\
         {/* Constraints for GPT-4 */} \\
         {\bf Category}: Mathematics and Reasoning $\rightarrow$ Word Problems; {\bf Language}: English; {\bf Style}: Roly-play; {\bf Attitude}: Command; {\bf Basic Question}: Ronnie was given 5 while Rissa was given thrice as much\ldots\\
         {\bf Conversation}: Human: Act as a helpful math assistant to answer this question: Ronnie \ldots \\
         \hline
    \end{tabular}
    \caption{Examples of prompt for GPT-4 for instruction data generation via in-context learning. The content generated by GPT-4 has been highlighted.}
    \label{tab:in-context_learning}
\end{table}

Unlike previous work, we focus only on tasks that may appear in real-world user instructions, as this represents what users genuinely want the \llms to achieve, providing a more practical evaluation of the instruction-following ability. Thus, we manually annotated the fine-grained task categories of the extracted instruction data from ShareGPT, primarily adhering to the traditional NLP task types commonly defined in previous research~\cite{flanv2,big-bench,llm_survey}.
For the convenience of conducting evaluations at different granularities, we design the categories as a multi-level tree structure, which facilitates efficient and in-depth analysis of the capabilities of \llms. Statistically, our category tree comprises 4 levels, with the first level containing {\bf 6} categories, the second level containing {\bf 25} categories, the third level containing {\bf 65} categories, and the fourth level containing {\bf 34} categories. We present the first and second level categories in Table~\ref{tab:category_tree}. 


As the goal of this work is to evaluate the performance of \llms on various instruction expressions, we also annotate the instruction style, attitude, and language for each instruction sample in the seed data, which are described as follows.

For instruction style, we specify the following five types:

\bi
\item {\bf Inquisitive} represents asking multiple questions on the same topic, or delving deeper into a particular question (See the first example in Figure~\ref{fig:sharegpt_examples}).
\item {\bf Reflective} represents asking multiple questions with the user's own thoughts and ideas (See the second example in Figure~\ref{fig:sharegpt_examples}).
\item {\bf Challenge} represents asking multiple questions, which are increasingly difficult (See the third example in Figure~\ref{fig:sharegpt_examples}).
\item {\bf Role-play} represents setting roles for both \llms and users, and conducting questioning under this setting. (See the fourth example in Figure~\ref{fig:sharegpt_examples}).
\item {\bf Concise} represents asking a question directly and clearly. (See the fifth example in Figure~\ref{fig:sharegpt_examples}).
\ei

For instruction attitude, we specify three types:
\bi
\item {\bf Polite} represents asking questions using gentle words, such as ``\emph{Could you answer the question\ldots}''.
\item {\bf Command} represents asking questions in a strong and imperative tone, such as ``\emph{Summarize this passage: \ldots}''.
\item {\bf Impatient} represents urging the \llm to respond to a certain aspect during the questioning process, such as ``\emph{Answer this question directly: \ldots, hurry up!}''
\ei

Moreover, for languages, we list all the languages included in the conversation, as users often switch between languages during the conversation. 

\subsection{Instruction Data Collection}
\label{sec:instruction-following_data_collection}

Generating diverse instructions is very challenging for human annotators, as it requires (1) the ability to transition between various instruction styles, attitudes, and languages, and (2) the capacity to produce a range of samples within a single category (\eg ``Grammar-based Rewriting''). Consequently, we propose employing the highly capable \llm, GPT-4~\cite{gpt4}, to simulate a variety of user types and generate diverse, high-quality instructions for each category. Please note that we do not directly incorporate instruction data from ShareGPT into our benchmark, because numerous \llms (\eg Vicuna~\cite{vicuna}, TÜLU~\cite{DBLP:journals/corr/abs-2306-04751}) have already utilized data from shareGPT for supervised instruction-tuning. Therefore, we only use ShareGPT data as seed data to guide GPT-4.  The data collection pipeline is depicted in Figure~\ref{fig:pipeline}.

For any leaf category (\eg ``Mathematics and Reasoning ''$\rightarrow$``Mathematical Calculations''$\rightarrow$ ``Algebraic Equation Problems''), we consider the following three steps to collect the instruction-following data.

In the first step, the goal is to generate constraints to guide GPT-4 to simulate specific user types, thereby preventing generation of unrealistic instructions. To achieve this, we treat the seed data as a sample pool and randomly select two samples as demos of in-context learning, each associated with a particular instruction style ($\mathcal{S}$), attitude ($\mathcal{A}$), and language ($\mathcal{L}$). We randomly sample target style, attitude, and language from these two demos to form constraints, compelling GPT-4 to learn from different instruction demos rather than excessively imitating one. For example, the constraints in Figure~\ref{fig:pipeline} is $\{\mathcal{S}=$``$Reflective$'', $ \mathcal{A}=$``$Polite$'', $\mathcal{L}=$``$English$''$\}$. Additionally, given that GPT-4 may struggle to generate high-quality mathematical or logical reasoning questions, we gather data from previous task-specific benchmarks as basic questions, which are then incorporated as part of the constraints. For example, we use the GSM8K~\cite{gsm8k} dataset as a basic question source for the ``Mathematics and Reasoning'' $\rightarrow$``Word Problems'' category. More details of the existing dataset resources included in \dataset are listed in Table~\ref{tab:basic_question_source}.
\begin{table}[]
    \centering
    \begin{tabular}{c|c}
         \hline
         {\bf Category} & {\bf Basic Question Source} \\
         \hline
         Word Problems & GSM8K~\cite{gsm8k}  \\
         Mathematical Theorem & ProofNet~\cite{proofnet}\\
         Combinatorics & Math~\cite{math} \\
         Numerical Calculation & Math~\cite{math} \\
         Common Sense Reasoning & StrategyQA~\cite{strategyqa} \\
         Logical Reasoning & LogiQA~\cite{logiqa} \\
         Text2Code & MBPP~\cite{mbpp} \\
         \hline
    \end{tabular}
    \caption{The basic question source of \dataset.}
    \label{tab:basic_question_source}
\end{table}

The goal of the second step is for GPT-4 to simulate a real-world user and generate high-quality instructions by adhering to the constraints. We use in-context learning to achieve this goal. As illustrated in Table~\ref{tab:in-context_learning}, we combine the task description, the two demos obtained from the first step, and the target constraints as input context for GPT-4. As demonstrated, GPT-4 learns different expressions from two demos and transforms the basic question into specific instruction ``\emph{Act as a helpful math assistant to \ldots }'' based on the {\bf Role-play} and {\bf Command} constraints. Following previous work~\cite{wiegreffe2021reframing,yuan2023distilling}, we adopt the {\it over-generate-then-filter} approach to obtain higher quality instructions. Thus, in this step, we prompt GPT-4 to make two predictions based on the same input, generating two instruction candidates.

In the third step, the objective is to select faithful and diverse instructions. We consider two selection methods, constraint-based pair-wise selection and similarity-based selection. Specifically, we first use GPT-4 to determine which of the two candidates adheres more closely to the constraints. We require GPT-4 to choose from three options, $\{first, second, tie\}$, and provide a rationale. Next, to ensure diversity, we calculate the similarity between the best candidate and the collected data in the dataset. Then, we only add the candidate to the dataset if the maximum ROUGE-L similarity is less than 0.6. 

\subsection{Dataset Analysis}
\label{sec:dataset_analysis}
\begin{table*}[h]
    \centering
    \begin{tabular}{c|c|c|c|c}
         \hline
         {\bf Category} & {\bf \#Tasks} & {\bf \#Samples} & {\bf \#Turns} & {\bf \#Input length} \\
         \hline
         Mathematics and Reasoning & 9 & 432 & 1.6 & 83.4 \\
         Language Understanding & 11 & 530 & 2.0 & 125.7 \\
         Language Generation & 14 & 730 & 1.9 & 157.1 \\
         Knowledge Utilization & 34 & 1596 & 2.6 & 72.7 \\
         Creation & 31 & 1498 & 1.9 & 63.4 \\
         Code & 5 & 240 & 2.1 & 105.5 \\
         \hline
    \end{tabular}
    \caption{The statistics of the \dataset dataset. `Category' represents the first-level category in the category tree. `\#Tasks' represents the number of tasks belonging to each first-level category. `\#Samples' represents the number of samples contained in each first-level category. `\#Turns' represents the average number of conversation turns included in each sample. `\#Input length' represents the average length of user input in each sample.
    }
    \label{tab:data_statistics}
\end{table*}

Table~\ref{tab:data_statistics} presents statistics of \dataset, which exhibits two main characteristics: (1) More fine-grained tasks are divided under each first-level category, such as ``Biology'' and ``Chemistry'' within the ``Knowledge Utilization'' category. (2) Each sample may comprise multiple turns of questions, simulating the process of human interaction with \llms.

\begin{figure}[h!]
  \includegraphics[width=\linewidth]{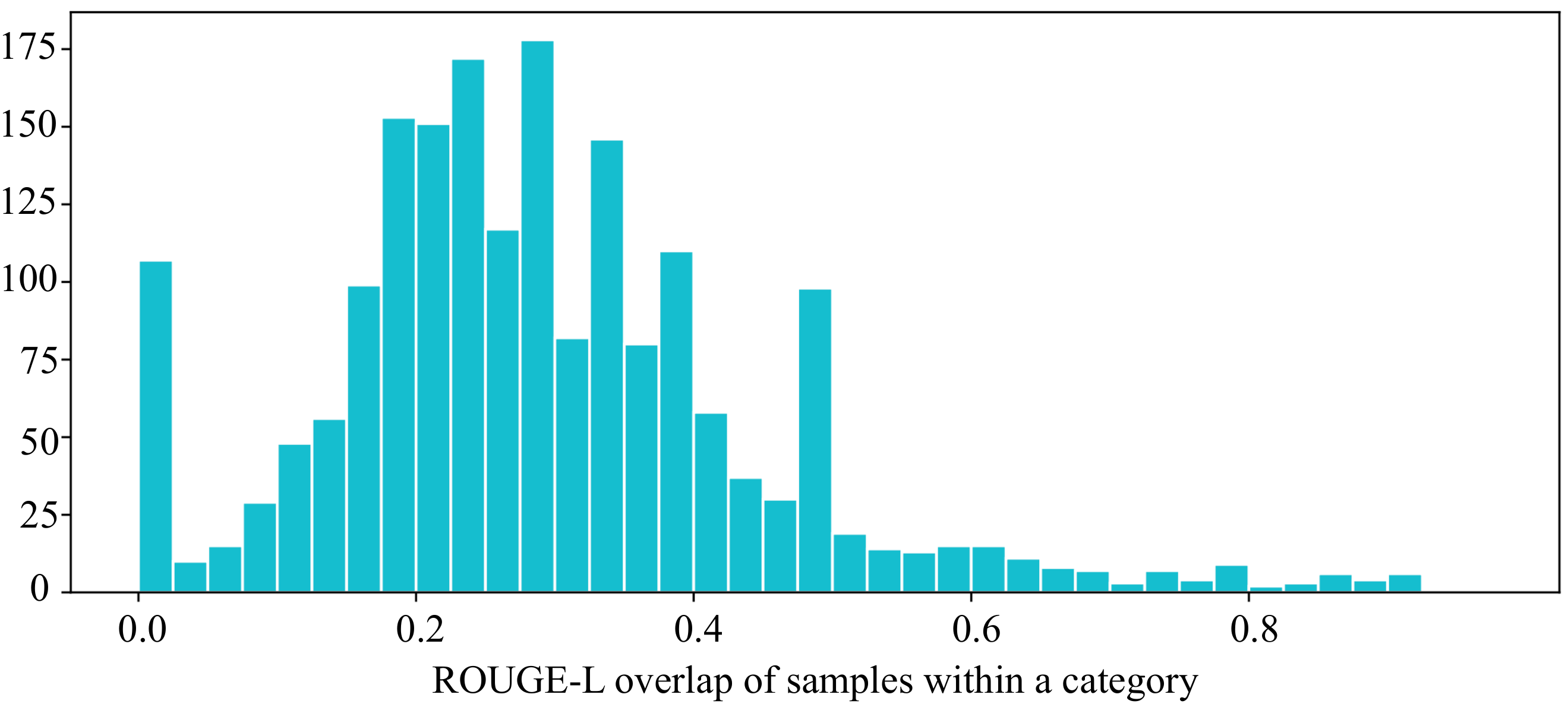}
\caption{Distribution of the ROUGE-L scores between instructions within a category.}
  \label{fig:similarity}
\end{figure}

To validate diversity within each category, we calculate the overlap degree of instructions in each category. Figure~\ref{fig:similarity} illustrates the similarity distribution of instructions.
For each instruction, we compute its highest ROUGE-L score with regard to other instructions in the same category. The results illustrate the diversity of instructions in \dataset.
\section{Experiments}
\label{sec:experiments}

\subsection{Experimental Setup}

\subsubsection{Baseline Models}
We select two representative types of \llms: (1) Pre-trained only \llms, including Llama~\cite{llama} and Llama2~\cite{llama2}; and (2) Instruction-tuned \llms, including vicuna-v1.3~\cite{vicuna}, vicuna-v1.5~\cite{vicuna}, Llama2-chat~\cite{llama2}. Considering that vicuna-v1.3 is instruction-tuned from Llama and vicuna-v1.5 is instruction-tuned from Llama2, we refer to vicuna-v1.3 as vicuna and vicuna-v1.5 as vicuna2 in this paper to make the notations consistent with Llama and Llama2.

\subsubsection{Evaluation Method}
We employ the {\tt LLM-as-a-judge} method to comprehensively evaluate \llm's responses~\cite{llm_judge}. {\tt LLM-as-a-judge} is a technique to score the performance of \llms by utilizing GPT-4. Researchers have discovered that GPT-4 can generate consistent scores and provide detailed justifications, which exhibit a high level of agreement with human experts. However, considering that GPT-4 has difficulty in accurately scoring math/code problems~\cite{cobbe2021training}, we include the standard answers for basic questions as a reference in the prompt given to GPT-4. Regarding the grading method, {\tt LLM-as-a-judge} considers two types, pair-wise comparison and single-answer grading. However, considering that we need to compare the performance of multiple \llms, we choose to use single-answer grading for more efficient evaluation. For different categories, we have manually annotated different scoring criteria to assist GPT-4 in generating scores that align with human preferences. For instance, in ``Mathematics and Reasoning'' tasks, the primary considerations include the clarity of steps, the correctness of reasoning, and the appropriateness of natural language explanations. Meanwhile, for ``Knowledge Unilization'' tasks, the primary considerations is on the adequacy of key points and whether the answers contain hallucination.

We explore the agreement between these two grading methods and human experts in Section~\ref{sec:Experimental_Results}.

\subsection{Experimental Results}
\label{sec:Experimental_Results}
\begin{figure*}[h!]
  \includegraphics[width=\linewidth]{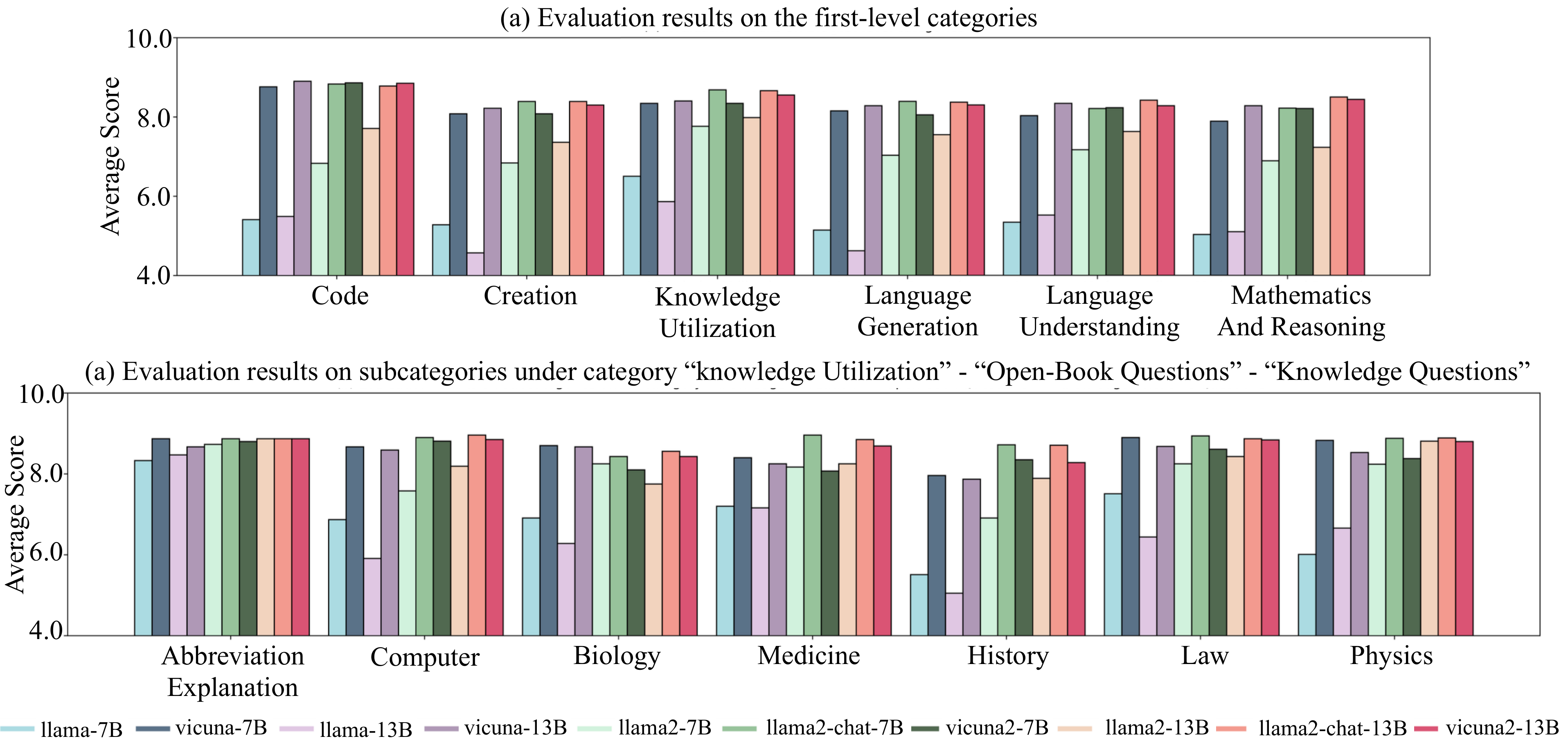}
\caption{Evaluation results of different \llms under different category granularity.}
  \label{fig:overall_performance}
\end{figure*}

\subsubsection{How do the existing \llms perform on \dataset?}
Figure~\ref{fig:overall_performance}-(a) shows the overall performance of ten \llms on the first-level categories of \dataset. First, comparing pre-trained \llms with instruction-tuned \llms, such as {\tt Llama-13B} and {\tt vicuna-13B}, we can see that instruction-tuning significantly impacts alignment with human preferences. Second, comparing different instruction-tuned \llms based on the same pre-trained \llms, such as {\tt vicuna2-7B} and {\tt Llama2-chat-7B}, we find that {\tt Llama2-chat-7B} has better instruction-following ability than {\tt Vicuna2-7B}. This is mainly because {\tt Llama2-chat-7B} utilizes an RLHF (reinforcement learning from human feedback) framework with two reward models for usefulness and safety to align with human preferences, enabling it to outperform the base \llm (\ie {\tt Llama2-7B}) under various user instructions. Finally, comparing \llms of different sizes indicates that increasing the model size significantly improves the instruction-following ability of the pre-trained \llms (such as {\tt Llama2-7B} and {\tt Llama2-13B}), but the impact on instruction-tuned \llms (such as {\tt Llama2-chat-7B} and {\tt Llama2-chat-13B}) is comparatively weaker.

\subsubsection{Can instruction-tuning consistently achieve stable improvements in more fine-grained categories?}
Figure~\ref{fig:overall_performance}-(b) illustrates the performance across all sub-categories under ``Knowledge Utilization''$\rightarrow$``Open-Book Questions''$\rightarrow$``Knowledge-Intensive Questions''. It can be seen that under a more fine-grained evaluation, the improvement brought by instruction-tuning is not consistent. For example, the instruction-following performance of {\tt vicuna2-7B} after instruction-tuning does not improve compared to its base \llm~{\tt Llama2-7B} in the two sub-categories: ``Biology'' and ``Medicine''.
This suggests that conducting a more fine-grained evaluation of \llms' instruction-following ability is necessary, as high scores in coarse categories (\eg ``Knowledge Utilization'') do not necessarily indicate stable performance in all finer sub-categories (\eg ``Biology''). Therefore, \dataset helps guide instruction-tuned \llms towards a more comprehensive direction for improvement, thus enhancing the interactive experiences for users with diverse requests.


\begin{table*}[]
    \centering
    \begin{tabular}{c|p{7.5cm}|p{7.5cm}}
         \hline
         {\bf From} & {\bf DINGO} & {\bf Basic Question}\\
         \hline
         Example-1 & User: Find out the number of ways Kim can place 10 identical lamps on 3 identical tables. Directly output the answer without any explanation. & User: Kim has 10 identical lamps and 3 identical tables. How many ways are therefor her to put all the lamps on the tables? \\
         & \llm: There are several ways to calculate the number of ways Kim can place 10 identical lamps on 3 identical tables … & \llm: First, let's count the ways to put 2 identical lamps on a single table. There are 2! ways to arrange … \\ 
         \hline
         Example-2 & User: For listed companies, … If the above statement is true, which statement cannot be true? A.A listed company needs financing and has no way to pay dividends. B … C … D … & User: For listed companies, … If the above statement is true, which statement cannot be true? A. A listed company needs financing and has no way to pay dividends. B … C … D … \\
         & \llm: Let’s analyze each statement one by one: … Therefore, the final answer is D. & \llm: Let's analyze each statement one by one: … Therefore, the final answer is D. \\
         & User: Please check each condition carefully! & \\
         & \llm: I apologize for any confusion … Therefore, the final answer is None of the above. & \\
         \hline
    \end{tabular}
    \caption{Two illustrative examples demonstrate why the \dataset is more challenging than basic questions in instruction following. We use ellipses to omit lengthy, less significant sentences.}
    \label{tab:examples_compare_with_basic_question}
\end{table*}

\subsubsection{Do diverse instruction types pose a challenge to \llms?}
\begin{figure}[h!]
  \includegraphics[width=\linewidth]{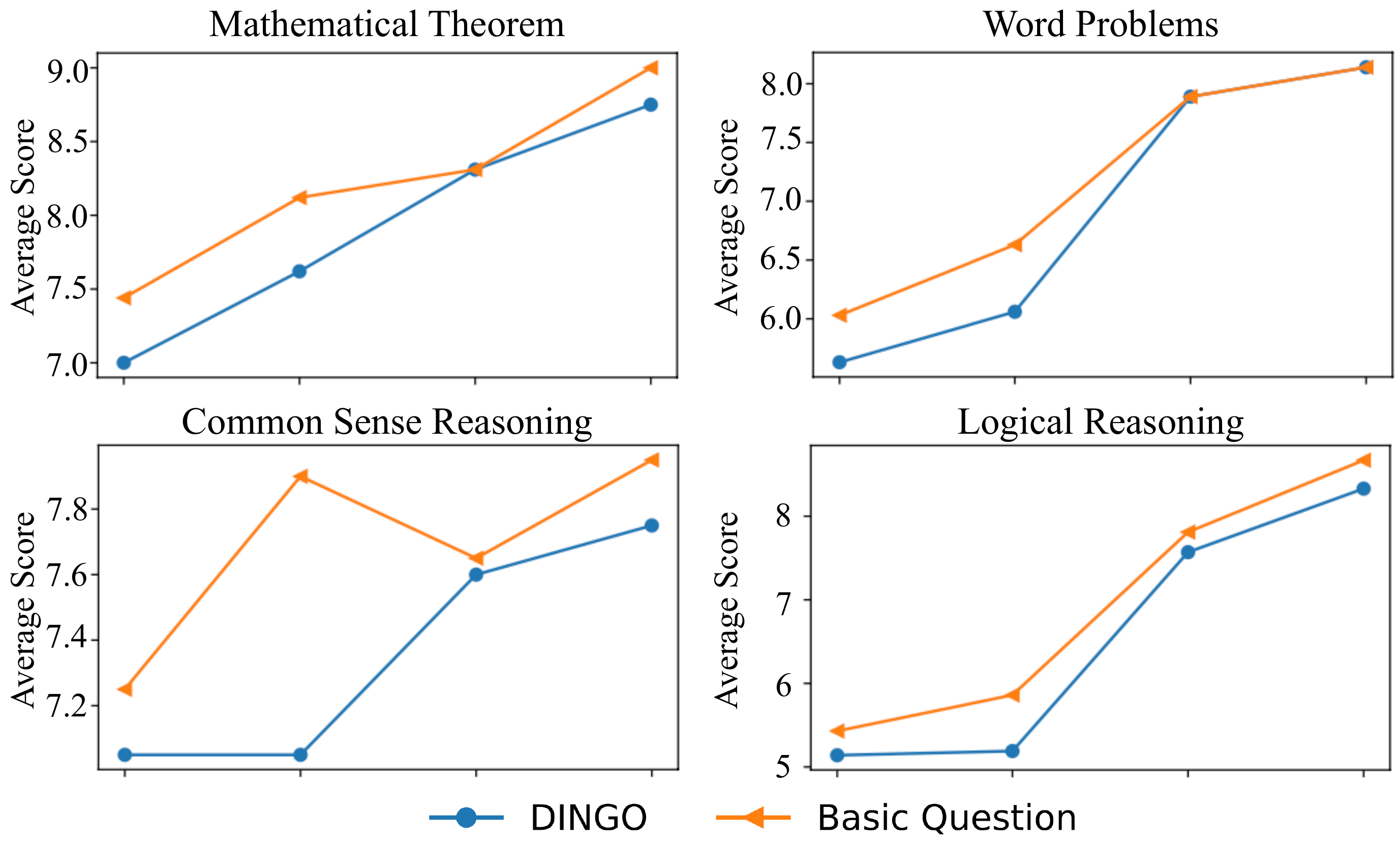}
\caption{Comparison of instruction following performance of \llms on \dataset and on basic questions.}
  \label{fig:compare_with_basic_question}
\end{figure}

 To investigate whether the diverse instruction types in \dataset present a significant challenge to \llms, we conduct an analysis on the categories containing basic questions. Specifically, we use four \llms to respond to basic questions and instructions in \dataset across four subcategories. The experimental results are shown in Figure~\ref{fig:compare_with_basic_question}. It can be observed that the instruction following scores of the four \llms on \dataset are lower than those on basic questions, indicating that the diverse instructions in \dataset are more challenging compared to standard questions. This also suggests that it is necessary to evaluate the \llms' instruction following ability using more diverse instructions, as an \llm may perform well in one mode of expression but not in others, implying that the \llm's robustness to diverse instructions in real-world scenarios might be insufficient.
 
 Additionally, to intuitively understand why the \llms perform poorly on diverse instructions, we present two examples in Table~\ref{tab:examples_compare_with_basic_question}. Example-1 indicates that when user instructions become more concise and require a concise output (\ie ``\emph{Directly output the answer without any explanation.}''), \llms still generate lengthy explanations that do not align with user instructions. Example-2 shows that when the instruction is in {\bf Challenge} style (\ie ``\emph{Please check each condition carefully!}''), the \llms may go against the original correct answer in order to cater to human users, \ie ``\emph{Therefore, the final answer is None of the above.}''.

\begin{figure}[]
  \centering
  \includegraphics[width=1.\linewidth]
  {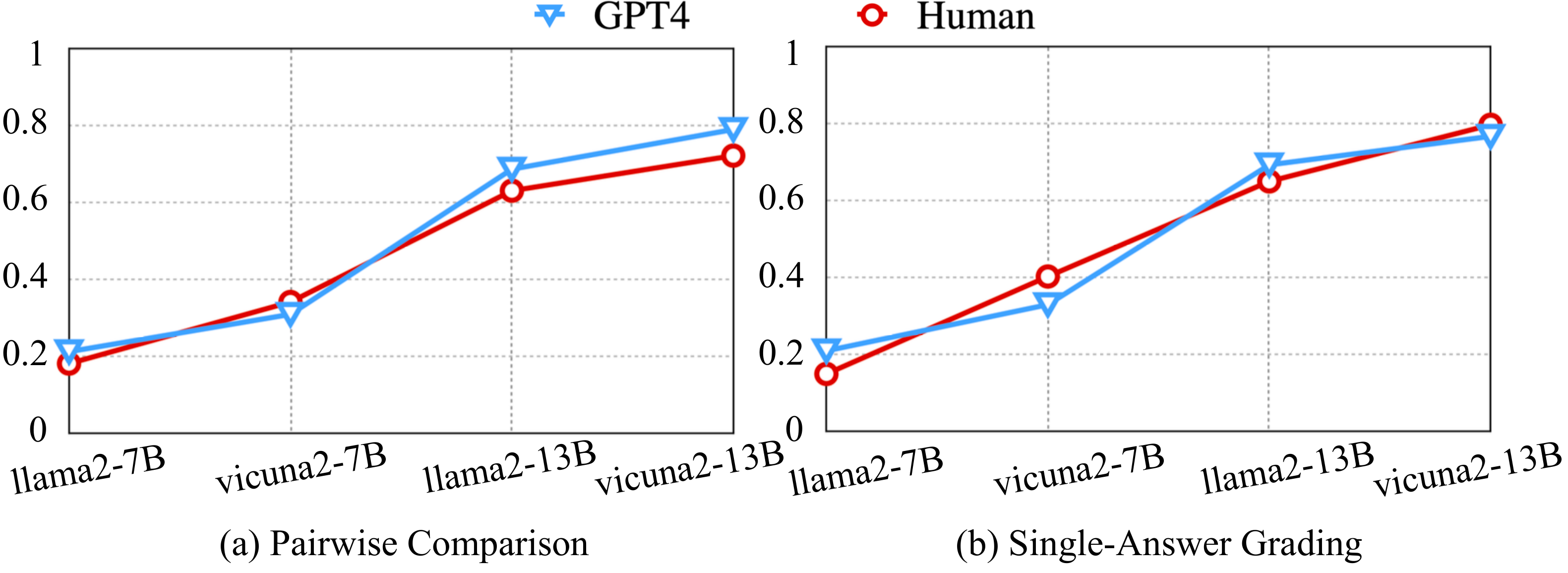}
\caption{Average win rate of four \llms under different judge methods.}
  \label{fig:human_gpt4_eval}
\end{figure}

\subsubsection{What is the agreement between human judge scores and GPT-4 judge scores?} To evaluate the agreement between GPT-4 and human experts, we choose 100 examples from \dataset and employ six human experts. Given a judge (\ie either GPT-4 or human expert), we ask the judge to score the responses of the \llms using two methods, (1) pairwise comparison and (2) single-answer grading. Pairwise comparison provides the judge a question and two potential answers, and asks the judge to decide which answer is more appropriate. Single-answer grading asks a juedge to assign a score to a specific answer. Figure~\ref{fig:human_gpt4_eval} shows the agreement between GPT-4 and humans under the two scoring methods. With pairwise comparison, GPT-4 has higher agreement with human. However, pairwise comparison would incur high cost. On the other hand, single-answer grading is more efficient. Thus, we recommend single-answer grading for rough identification of model issues, and pairwise comparison for more detailed evaluations.

\section{Related Work: Evaluation of \llms}
\label{sec:related_work}
For benchmarking the effectiveness of \llms, various evaluation frameworks have emerged. 
Frameworks such as HELM~\cite{helm} and BIG-BENCH~\cite{big-bench} focus on the effectiveness of \llms on a wide range of NLP tasks, mainly evaluating the problem solving ability of the model, without paying attention to the \llm's instruction-following ability. Recently, some work has started to focus on the instruction-following ability of \llms. For example, InstructEval~\cite{chia2023instructeval} focuses on evaluating the ability of Instruction-Tuned \llms on three aspects, including problem solving, writing, and alignment. Alpaca Farm~\cite{dubois2023alpacafarm} and Chatbot Arena~\cite{llm_judge} focus on evaluating the open-ended instruction-following ability of \llms. However, there are two main differences between \dataset and the above studies: (1) a diverse set of instructions based on real-world scenarios, which can comprehensively evaluate the model's instruction-following performance. (2) a fine-grained task category tree, which can deeply analyze \llm's instruction-following ability on fine-grained task types and pinpoint the deficiencies for further improvement.
\section{Conclusion}
\label{sec:conclusion}
In this paper, we have presented a diverse and fine-grained instruction-following evaluation dataset \dataset. Based on a multi-level category tree with 130 nodes derived from real-world user requests, \dataset includes 5026 diverse instructions. Our experiments demonstrate that (1) while an instruction-tuned \llm may excel in broad categories, its performance can vary in fine-grained categories; (2) diverse instructions pose greater challenges for \llms to generate responses that match human preferences. 

\section*{Acknowledgments}
This paper is supported by the Science and Technology Development Fund of Macau SAR (File no. 0081/2022/A2, 0123/2022/AFJ, and 0015/2019/AKP), and GuangDong Basic and Applied Basic Research Foundation (No. 2020B1515130004). This paper is also partly supported by the NSF of China (62122090, 62072461 and 62072458), the Fund for Building World-Class Universities (Disciplines) of Renmin University of China, the Beijing Natural Science Foundation (L222006), and the Research Funds of Renmin University of China.
\bibliography{aaai24}
\end{document}